\documentclass{isprs} 
\usepackage{subfigure}
\usepackage{setspace}
\usepackage{geometry} 
\usepackage{siunitx} 
\usepackage{epstopdf}
\usepackage[labelsep=period]{caption}  
\usepackage[british]{babel} 
\usepackage[hang]{footmisc}
\usepackage{amsmath}
\usepackage{amssymb}
\usepackage{booktabs}
\usepackage{url}
\newcommand{\AVL}{$\mathrm{egenioussBench}$}

\begin{document}

\title{\AVL: A New Dataset for Geospatial Visual Localisation}



\author{
Phillipp Fanta-Jende\textsuperscript{1},
Francesco Vultaggio\textsuperscript{1,2}, Alexander Kern\textsuperscript{3},  Yasmin Loeper\textsuperscript{2},  Markus Gerke\textsuperscript{2}
}

\address{
   \textsuperscript{1 } Unit Assistive and Autonomous Systems, Center for Vision, Automation and Control,\\ AIT Austrian Institute of Technology, Vienna, Austria - (phillipp.fanta-jende, francesco.vultaggio)@ait.ac.at\\
\textsuperscript{2 }Institute of Geodesy and Photogrammetry, Technische Universität Braunschweig,\\Brunswick, Germany - (m.gerke, y.loeper)@tu-braunschweig.de\\
   \textsuperscript{3 }Institute of Flight Guidance, Technische Universität Braunschweig, \\Brunswick, Germany - a.kern@tu-braunschweig.de
}



\abstract{

We present \AVL, a visual localisation benchmark built on geospatial reference data: a city-scale airborne 3D mesh and a CityGML LoD2 model. This pairing reflects deployable mapping assets and supports true scalability beyond traditional SfM-based approaches. The query data comprise smartphone images with centimetre-accurate, map-independent ground truth obtained via PPK and GCP/CP-aided adjustment. From 2{,}709 images, we derive a non-co-visible subset by estimating the full co-visibility matrix from rendered depth and selecting a maximum independent set; the released data include a test split of 42 non-co-visible images with withheld ground truth and a validation split of 412 sequential images with poses, e.g.\ for training of pose regressors and self-validation. The benchmark features a public leaderboard evaluated with binning metrics at multiple pose-error thresholds alongside global statistics (median, RMSE, outlier ratio), ensuring fair, like-for-like comparison across mesh- and LoD2-based methods. Together, these design choices expose realistic cross-view and cross-domain challenges while providing a rigorous, scalable path for advancing large-scale visual localisation. We make the evaluation code and data availeable at \url{https://github.com/fratopa/egenioussBench} and \url{https://www.egeniouss.eu/}

}

\maketitle

\section{Introduction}\label{Intro}
 \sloppy

Visual localisation - estimating a camera’s position and orientation from images - underpins a wide variety of platforms and applications such as autonomous driving~\cite{sattlerBenchmarking6DOFOutdoor2018}, robotics~\cite{maggio2023locnerf}, Unmanned Aerial Vehicles (UAVs)~\cite{Wu.2024}, and augmented reality~\cite{pang2023ubipose}.
It is especially critical when GNSS is unavailable or unreliable, for instance indoors, in urban canyons, or under jamming and spoofing. 

Traditional approaches rely on Structure-from-Motion (SfM) maps~\cite{sattlerEfficientEffectivePrioritized2017,sarlin2019coarse}, but these suffer from limited scalability and heavy storage demands, often reaching hundreds of gigabytes~\cite{trujilloEfficientSceneCompression2020a}. Deep learning alternatives such as scene coordinate regression~\cite{wangGLACEGlobalLocal2024} offer efficiency and speed, yet still struggle with large-scale deployment. Moreover, most localisation techniques assume access to ground imagery which poses a constraint to their ability to scale to large scenes. These limitations motivate the need for alternative reference representations.

3D meshes are a promising option~\cite{panekMeshLocMeshBasedVisual2022a,brachmannAcceleratedCoordinateEncoding2023a}. Unlike SfM point clouds, they are naturally smaller in size, allow rendering from arbitrary viewpoints, enabling cross-view matching between ground-level imagery and aerial reconstructions~\cite{vultaggio_et_al_lc3d2024}. This is vital for scalable localisation, particularly as municipalities increasingly provide detailed meshes and city models~\cite{syedabdulrahmanDigitalLandscapeSmart2024}.

\begin{figure}
    \centering
    \includegraphics[width = 0.8\linewidth]{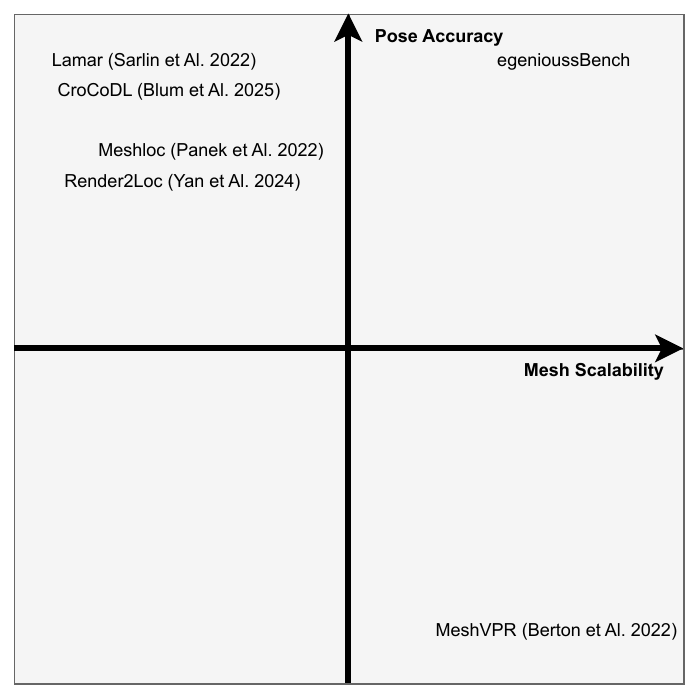}
    \caption{Comparison of egenioussBench to other state of the art mesh-based visual localisation datasets}
    \label{fig:comparison}
\end{figure}

Progress in this area has been limited by the nature of available datasets. Existing approaches have typically relied on two types of data: large-scale meshes with coarse geometry and imprecise ground-truth poses for the query images~\cite{bertonMeshVPRCitywideVisual2024a}, or smaller datasets where accurate ground-truth poses are obtained by co-registering query images with dense, high-resolution meshes reconstructed from street-level imagery~\cite{panekMeshLocMeshBasedVisual2022a,sarlin2022lamar} (see Figure~\ref{fig:comparison}).

While low-resolution datasets are valuable for training Visual Place Recognition (VPR) models, their limited accuracy makes them unsuitable for evaluating visual localisation performance. Conversely, datasets that depend on co-referencing with high-resolution 3D models inherently require meshes of very fine detail to produce precise ground-truth poses. This dependence not only makes such datasets impractical for large-scale mapping applications, but it also renders the localisation task very easy and will lead to unrealistically accurate results. 

In this work, we contribute a new dataset for mesh-based visual localisation with centimeter-accurate ground truth collected in a map-independent fashion, thus allowing us to pair ground-level query poses with a challenging aerial map. 

In cases where a mesh might not be available an additional source for geospatial reference is 3D city models, commonly provided as City Geography Markup Language (CityGML) models\footnote{https://www.ogc.org/standard/citygml/}. They are freely available for many cities, do not require a large amount of memory, and the scene re\-pre\-sen\-ta\-tion database does not have to be generated from images. However, since CityGML models consist purely of potentially low-detailed geometry compared to the mentioned textured mesh data, it represents a challenge for traditional feature-based localisation techniques. Simplified meshes without texture and general Level of Detail (LoD) meshes are comparable to CityGML models. Current work addresses the challenging yet promising reference data type~\cite{Zhu.2024,zhu2025lodlocv2aerialvisual,Panek.2023}. As with the work on large-scale meshes, the lack of precise and independent ground truth for the query data and georeferenced 3D models often presents an obstacle to evaluating visual localisation techniques. 

Our contributions are threefold: (i) we introduce \AVL, the first benchmark coupling a high-resolution aerial 3D mesh with a CityGML LoD2 model; (ii) we release map-independent ground-level smartphone imagery with centimetre-accurate ground truth poses for cross-view evaluation; and (iii) we establish a benchmark protocol and leader-board to foster rigorous comparison of mesh- and object-based localisation methods. The dataset can be found at \url{https://www.egeniouss.eu/}

\section{Related Work}\label{RelWork}

\subsection{Mesh-based Visual Localisation}
Mesh-based visual localisation techniques can be broadly divided into two categories: iterative approaches and hierarchical ones. Iterative approaches assume to have access to an initial position and then adopt a render and compare strategy \cite{cstar,yanRenderandCompareCrossview6DoF2023} to refine the pose estimate of the query image. Instead, hierarchical approaches \cite{panekMeshLocMeshBasedVisual2022a,vultaggio_et_al_lc3d2024} employ multi-stage retrieval methods to narrow down plausible images to a subset of candidate views and then determine the query position through accurate matches against these.

An often unspoken assumption of mesh-based localisation techniques is that the poses from which the reference images were captured, $P_{\text{ref}}$, are good poses from which to render images to be used in the localisation pipeline \cite{panekMeshLocMeshBasedVisual2022a}. While this is a sound assumption for 3D data collected from ground level, this assumption breaks in the case of an aerial dataset, where the large viewpoint difference makes reference poses harder -- if not impossible -- to localise from \cite{yanRenderandCompareCrossview6DoF2023}.

Another assumption in the mesh-based localisation literature is that a render-and-compare approach will naturally converge to the correct position if initialised in the neighbourhood of the ground-truth query pose, $P_{\text{gt}}$ \cite{cstar,yanRenderandCompareCrossview6DoF2023}. This assumption is grounded in the idea that real-world imagery is often trivially matched if captured from the same pose, defining an error cost function -- $\mathcal{E}(P_r)$ -- of matching a query image with a pose rendered at pose $P_r$, 

$$\arg\min_{P} \mathcal{E}(P_r) = P_{\text{gt}}$$

This is not always the case: As shown in \cite{vultaggio_et_al_lc3d2024}, matching query images against renderings from the ground-truth query poses can often lead to substantial pose errors due to errors in the mesh reconstruction process, such as obstructions, blind spots, or inaccurate correspondences (see Figure~\ref {fig:data_overview_query}). 

With this work, we aim to challenge the assumptions underlying current mesh-based localisation techniques and highlight new areas for improvement, such as viewpoint selection, which has been recognised as a crucial aspect of localisation in the robotics community \cite{digiammarinoLearningWhereLook2025} but which is often overlooked in the visual localisation field.

\subsection{Mesh-based Localisation Datasets}

Most localisation datasets consist of posed reference images, posed query images, and a 3D model. Often these datasets assume the 3D model to take the form of a SfM point cloud in which each point is associated with a local visual descriptor. Among these, the popular ones are Aachen \cite{aachen11}, 12 scenes \cite{12scenes}, inlooc\cite{inloc}, in these we find represented indoor \cite{inloc,12scenes} outdoor \cite{aachen11}, and challenging queries collected at different times of day and weather conditions.

Mesh-based localisation was first explored in indoor settings by \cite{cstar} as a loop closure and drift-correction technique. Full visual localisation using mesh maps was first proposed in MeshLoc~\cite{panekMeshLocMeshBasedVisual2022a}. Here, the retrieval stage was conducted on real images, and the local matches and 3D lifting were performed on synthetic RGB and depth views. 
Other prominent datasets are Lamar ~\cite{sarlin2022lamar} and CroCoDL \cite{blumCroCoDLCrossdeviceCollaborative2025}, which provide high-resolution 3D meshes collected from a custom mobile mapping setup and generate the ground truth of the query images by co-registering them with the 3D map. Both MeshLoc, Lamar, and CroCoDL provide meshes with high-resolution textures and fine-grained geometry, made possible by careful ground-level data collection. 

In this work, we argue that such datasets, while valuable, provide an unrealistically good reference mesh, which, while possible to obtain for small scenes, becomes prohibitively expensive to acquire for city-scale applications.

A dataset providing true city-scale data is MeshVPR~\cite{bertonMeshVPRCitywideVisual2024a}, however here the query pose ground truth position has not been estimated to a level of precision sufficient for visual localisation as this dataset was originally developed to train Visual Place Recognition (VPR) models. Render2Loc \cite{yanRenderandCompareCrossview6DoF2023} provides drone-captured data and accurate ground truth obtained through co-referencing of query images and a 3D model, however is not representative of true large-scale data collection, having been collected from a low-flying drone resulting in an estimated GSD of 1.03cm.

Another dataset with a similar ground truth query pose acquisition process to ours is the SLAM dataset recently released by \cite{krishnanBenchmarkingEgocentricVisualInertial2025}, here the query pose is also obtained through the use of GCPs and image reconstruction, but being a SLAM dataset, it does not provide a reference map for visual localisation.

This work aims to fill a gap in datasets currently released by providing a true city-scale mesh dataset collected from airborne imagery and centimetre-accurate ground-level imagery. Both datasets underwent a professional photogrammetric bundle adjustment involving GCPs collected in the scene. 

\subsection{Object-based Visual Localisation}
Object-based visual localisation uses an object's geometric characteristics as features to be matched in the query pose estimation process. The geometric features can be utilised in their original form, e.g., through projections or through rendered representations. The object data consists of simplified, textureless, and low-detail 3D models.

The approaches for the object data are similar to those that use 3D meshes as reference data. 
In hierarchical approaches, the 3D model is usually used to lift the 2D-2D feature matches of the query image with the retrieved reference images with known poses from the database into 3D~\cite{panekMeshLocMeshBasedVisual2022a}. In the absence of reference images for image retrieval, sampling approaches are used to generate synthetic images that utilize the object data. The 2D-2D feature matching is then based on synthetic images instead of reference images~\cite{Panek.2023}. 

In iterative pose estimation approaches, the correspondence between the features from the query image and the features from renderings of pose hypotheses is calculated and optimised~\cite{Zhu.2024,zhu2025lodlocv2aerialvisual,loeper_et_al_lc3d2024}. 
A common problem with the use of synthetic images is the feature matching or the calculation of correspondence between the query image and the non-photorealistic renderings. Previous work has shown that cross-domain matching can be successful~\cite{Tomesek.2022,Brejcha.2020,MikolkaFlory.2022}. Nevertheless, there is a need for further development to calculate the correspondence between the query image and renderings of textureless and low-detailed geometry models, e.g. by developing feature matching approaches adapted to textureless 3D models.

Instead of explicitly using 3D models, other approaches implicitly use the models to train a neural network that predicts the pose~\cite{Acharya.2022,Yao.2024}. 
In contrast to pose regressors that depend on SfM methods to generate training data or require images with ground truth poses, these approaches use synthetic images or features from synthetic images to train the neural network. 
Specifically, the synthetic image is rendered using 3D models and the ground truth query image poses. The features are then generated from the synthetic image, e.g. in the form of edge and segmentation maps. However, it must be emphasised that not only textureless and low-detailed models are used as training data. In addition, textureless models were either textured or rendered photorealistically before the edge and segmentation maps were generated. 
To date, research on using low-detailed, textureless models for pose regression remains limited.

\subsection{Object-based Visual Localisation Datasets}
With CADLoc~\cite{Panek.2023}, the authors provide a benchmark dataset for analysing imperfect 3D models from the Internet for use in visual localisation. In addition to the models, the dataset also contains query and reference images. Since the scale of the scenes is not known, the usual metrics of visual localisation cannot be used for evaluation. Instead, the mean and maximum Dense Re-Projection Error~\cite{Wald.2020} are used. 

For the visual localisation of UAVs, LoD-Loc provides two datasets consisting of LoD models and query images~\cite{Zhu.2024}.
The query images of the datasets from LoD-Loc originate from the UAVD4L dataset~\cite{Wu.2024} and from CrossLoc~\cite{Yan.2021}. The UAVD4L dataset contains rendered images, query images, their ground truth and digital surface model (DSM). The LoD model was generated in LoD3 from the oblique images of the UAVD4L dataset~\cite{Zhu.2024}.
The LoD model for the query images from CrossLoc is in LoD2 and from the Swiss federal authorities. 

Object-based visual localisation has not yet been extensively researched. 
In particular, there is a lack of datasets that contain both query images with precise, map-independent ground truth poses and georeferenced 3D models. However, LoD models are already available for many cities worldwide. Our contribution is to provide these two components in one dataset.

\section{The Benchmark Challenge}
\AVL\ evaluates state-of-the-art visual localisation under realistic, city-scale conditions. Participants are provided with two forms of geospatial reference data - a high-resolution aerial mesh and a CityGML LoD2 model - together with query images captured using a smartphone.

\begin{figure}[htp]
    \centering
    \includegraphics[width=0.75\linewidth]{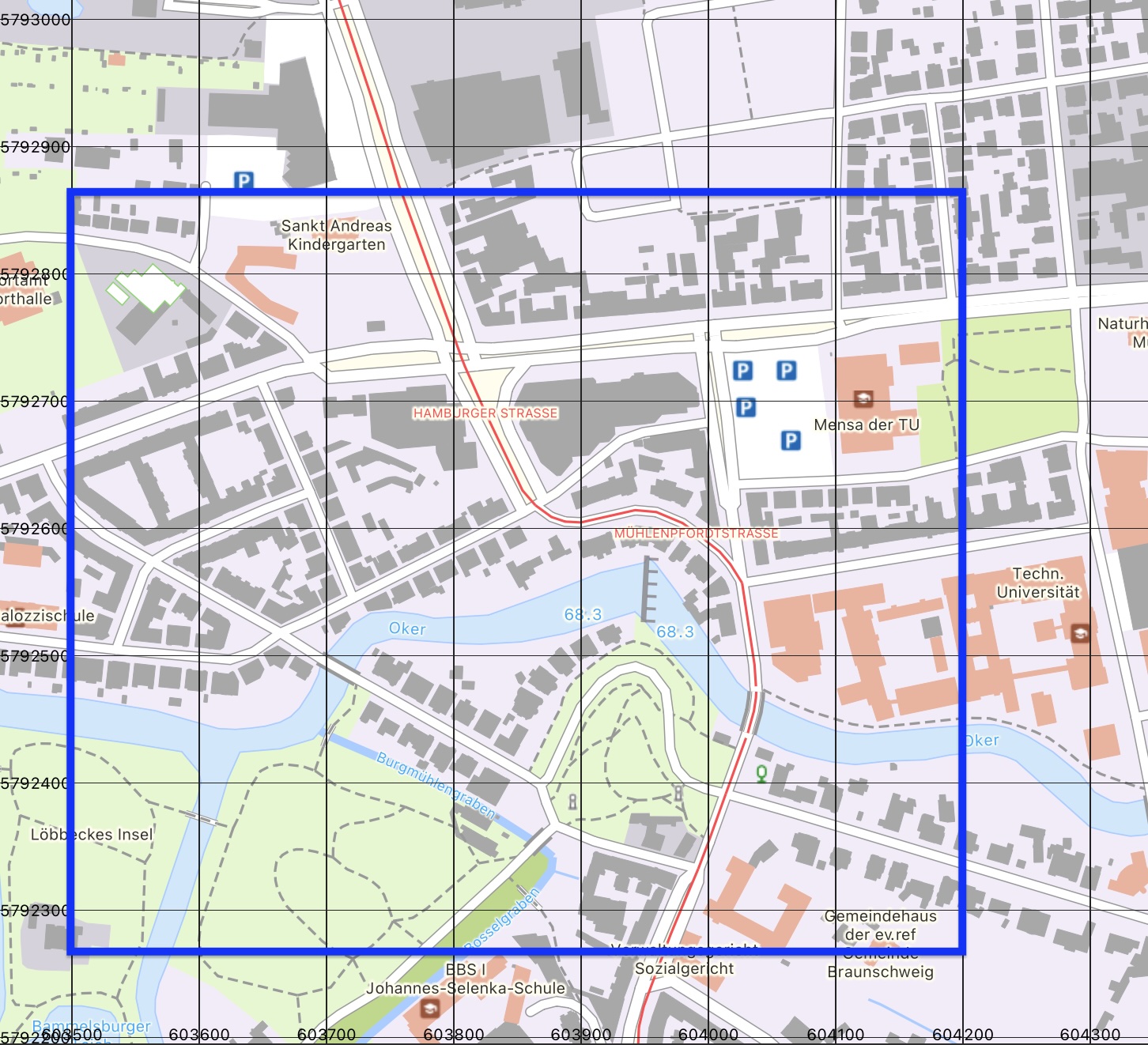}
    \caption{Overview of the area of interest in Braunschweig. Map source: basemap.de}
    \label{fig:Map}
\end{figure}

Unlike previous benchmarks, the query images are non-overlapping in object space. This design prevents the use of multi-view geometry from image sequences and enforces true cold-start localisation. Methods must therefore rely either on implicit regressors (e.g. learning-based global pose estimation) or on explicit matching against the provided references (mesh or LoD2).

The query set is divided into two partitions: one with full ground-truth poses for training and self-validation, and one with only approximate poses. The withheld ground-truth of the second partition is used for evaluation, with results published on a public leaderboard.

Participants will be asked to submit a CSV file with full 6DOF pose information per image. We will then evaluate each submission by binning according to thresholds compared to the ground truth (0.5m,2°/2m,5°/5m,10°), and overall median translation and rotation errors.

To ensure fair comparison, the challenge is evaluated se\-pa\-ra\-te\-ly for mesh-based and LoD2-based localisation. Examples of possible approaches are given in~\cite{vultaggio_et_al_lc3d2024} and  \cite{loeper_et_al_lc3d2024}.

We aim to organise a scientific workshop or special issue featuring the results submitted by participants.

\section{Datasets}
\AVL\ covers an area within the city of Braun\-schweig, Germany, in Fig.\ref{fig:Map}, a map is shown. The data is projected in UTM, zone 32N, and spans 570m in N-S and 700m in E-W direction. Corners at North West: [N: 5792850m, E: 603500m] and South East: [N: 5792280m, E: 604200m]. 

\begin{figure}[h]
    \centering
\includegraphics[width=0.9\linewidth]{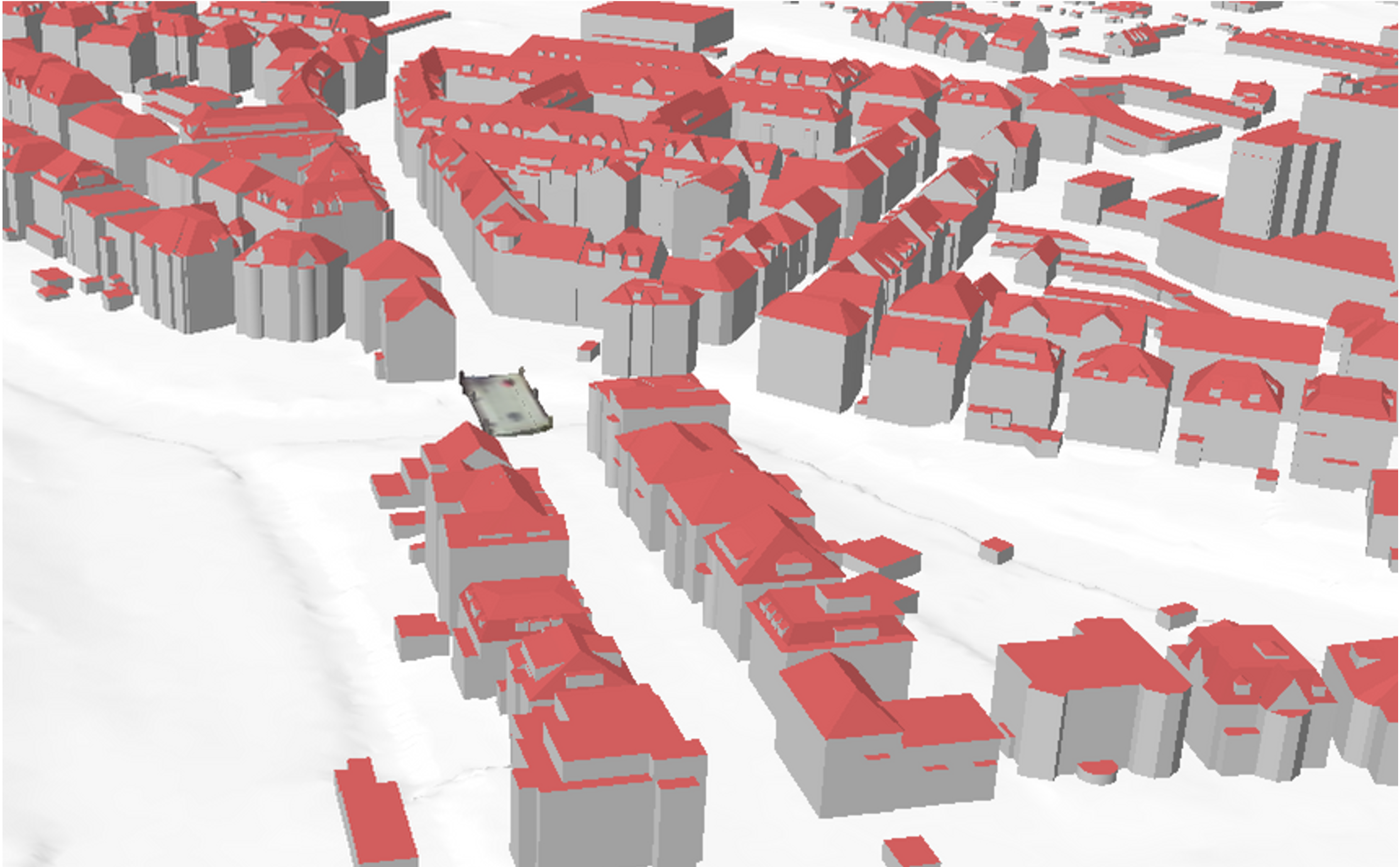}
    \caption{LoD2 model; Data: City of Braunschweig}
    \label{fig:data_overview_LOD}  
\end{figure}

\begin{figure}[h]
    \centering
\includegraphics[width=0.99\linewidth]{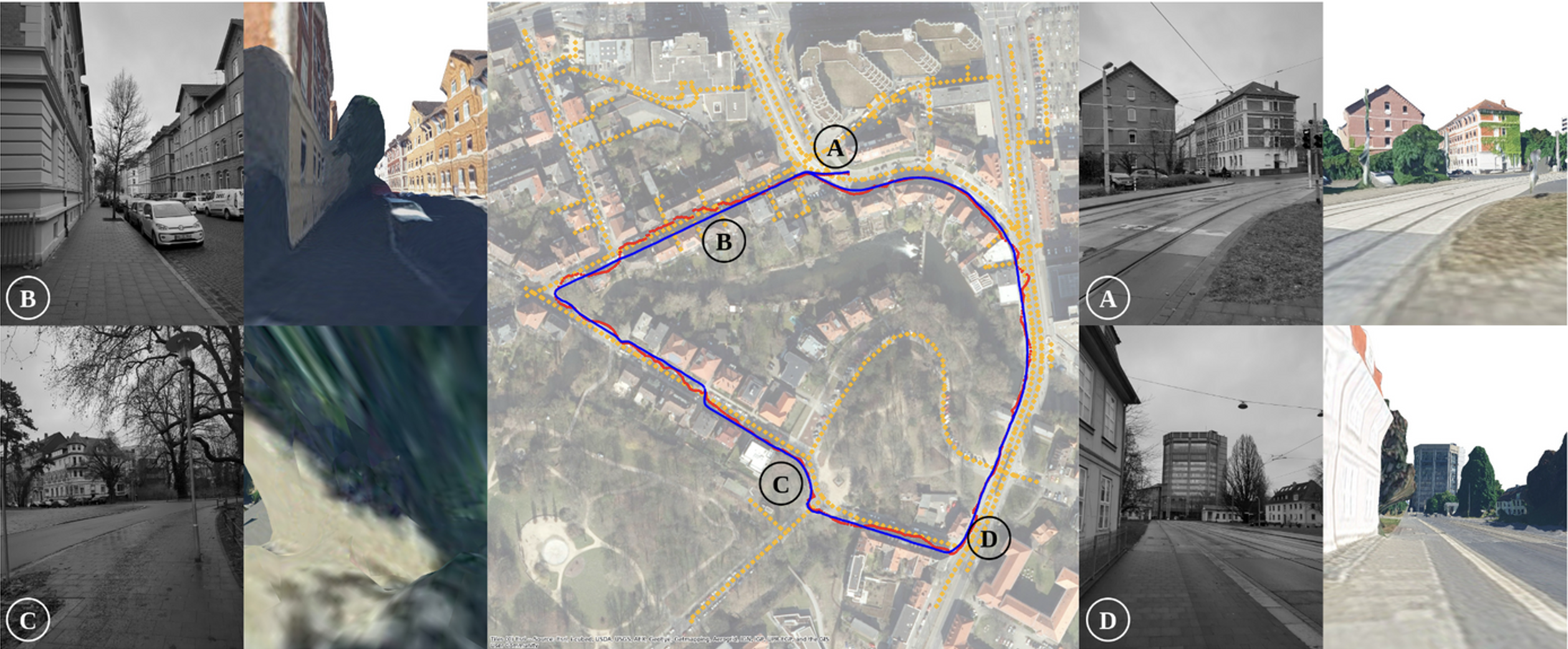}
    \caption{Left and right: examples of the query
images and the view rendered from its reference
Center: reference trajectory of the camera in blue, the smartphone’s internal GNSS pose estimate in red, and the sampled views in orange, source: (Vultaggio et al., 2024).}
    \label{fig:data_overview_query}  
\end{figure}

The dataset consists of three components:
\begin{enumerate}
    \item Airborne-image-based 3D mesh (oblique imagery, 7.5 cm GSD) - reference for mesh-based localisation (Fig.~\ref{fig:data_overview_mesh});
    \item CityGML LoD2 model - reference for object-based localisation (Fig.~\ref{fig:data_overview_LOD});
    \item Smartphone query images — ground-level queries, with cm-level ground-truth poses (Fig.~\ref{fig:data_overview_query}).
\end{enumerate}

\subsection{Airborne-image-based 3D mesh}
\begin{figure}[h]
    \centering
\includegraphics[width=0.9\linewidth]{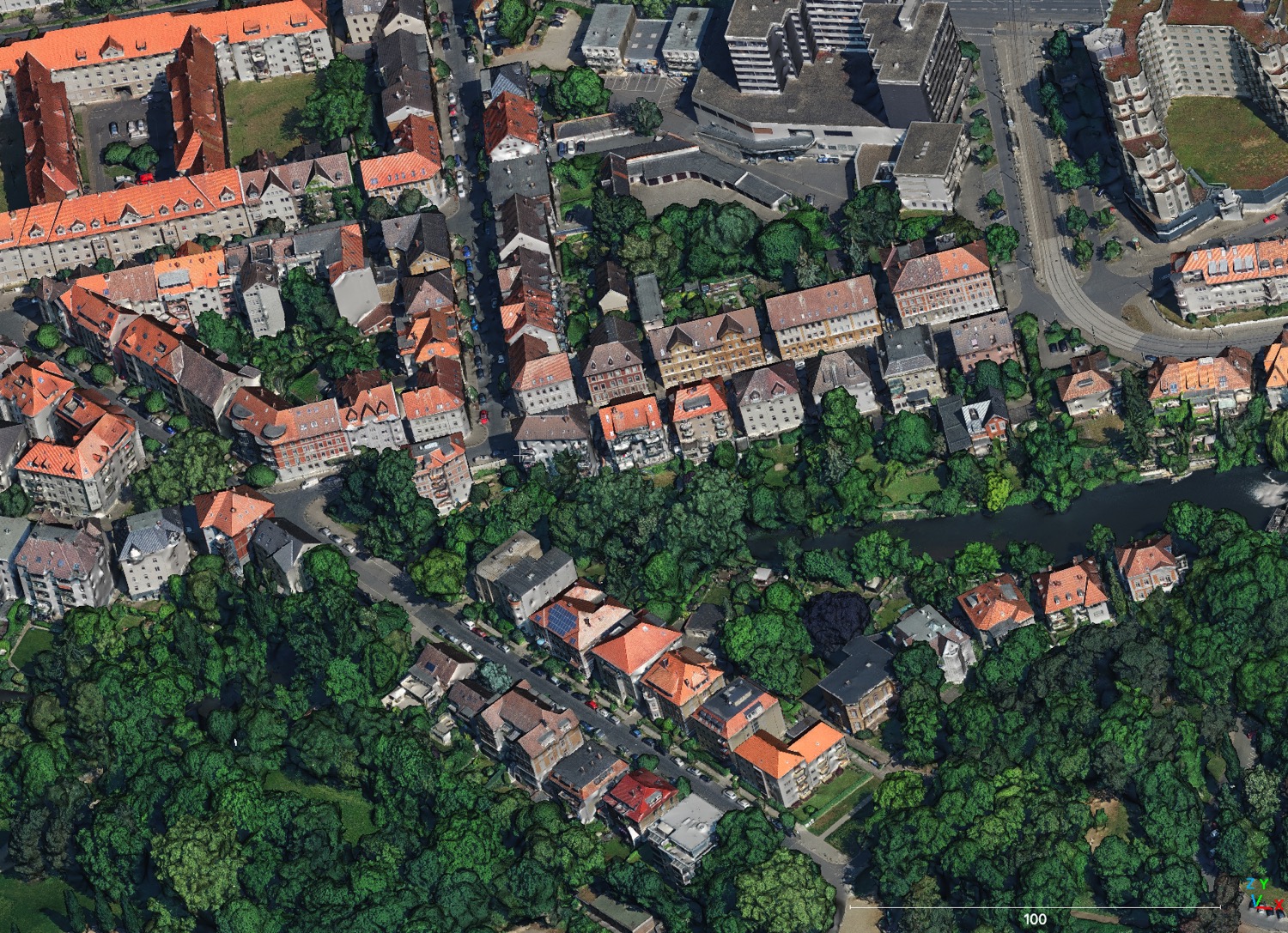}
\includegraphics[width=0.9\linewidth]{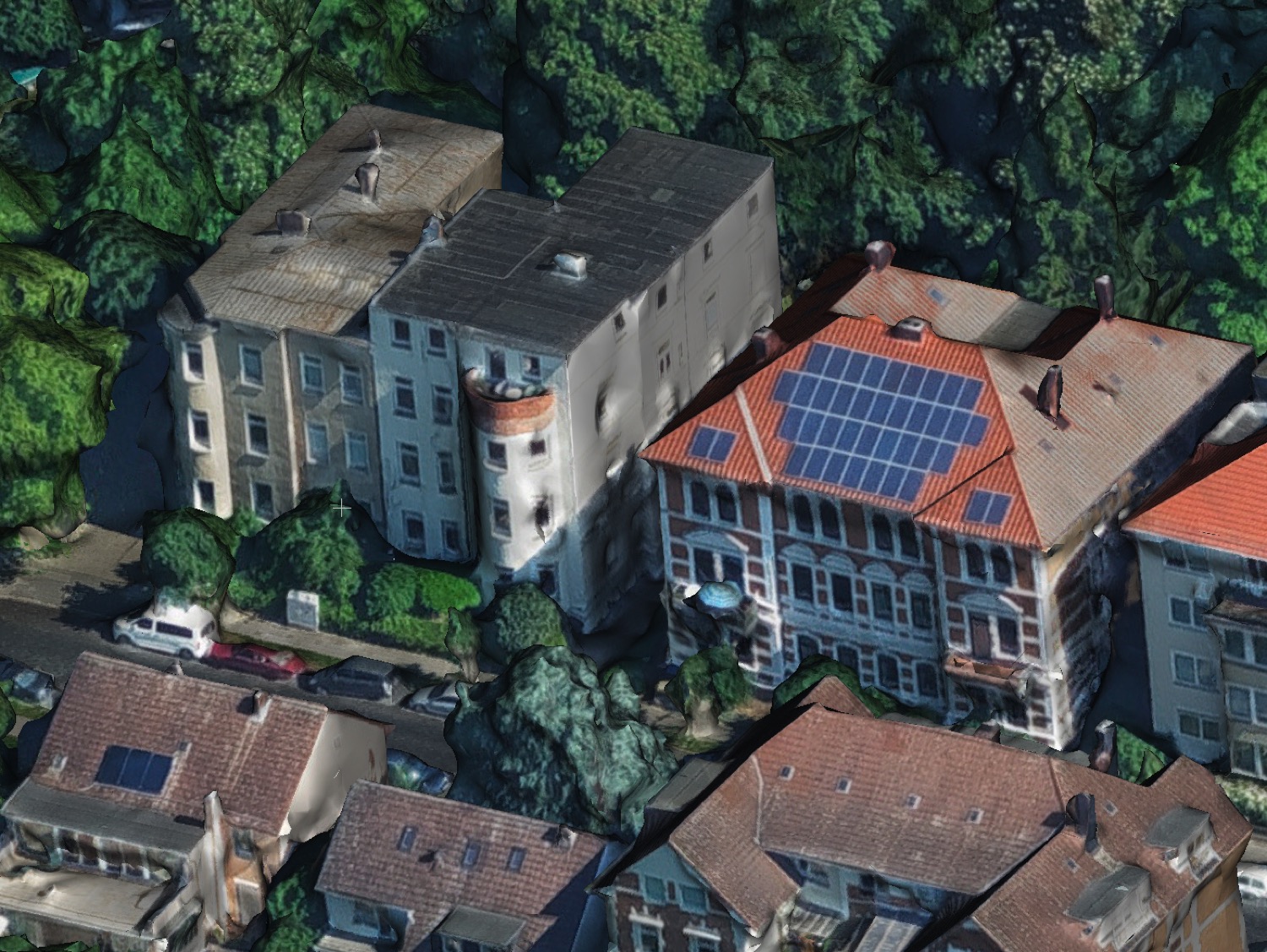}
    \caption{Geospatial mesh data; Data: Geofly, Processing: Skyline}
    \label{fig:data_overview_mesh}  
\end{figure}

Airborne oblique imagery acquired in May 2023 served as the basis for the 3D mesh reconstruction, see Fig.~\ref{fig:data_overview_mesh}.

\begin{tabular}{lp{5.4cm}}
Camera system & UltraCam Osprey 4.1 \\
Flying height (AGL) & $\sim$1550 m \\
GSD (nadir / oblique) & 7.5 cm / 6.5 cm (centre) \\
Georeferencing accuracy & $\sim$1 GSD (XY), 1.5 GSD (Z) \\
\end{tabular}

\subsection{CityGML LoD2} 
The second reference dataset is provided in the form of the 3D city model of Braunschweig~\cite{lgln.2024}, see Fig.~\ref{fig:data_overview_LOD}. The 3D city model is given in LoD2. The dataset is supplied in OGC Standard CityGML format. A standardised process is used to create the 3D building models of Braunschweig. The process is based on the building outlines from cadastral maps, the DTM with \SI{5}{\m} grid resolution and 3D measurement data from laser scanning or matching point cloud~\cite{lgln.2024}. 

The fundamentals for creating 3D building models are defined by product and quality standards~\cite{AdV.2025}. In accordance with these standards, the building footprints are extracted from the cadastre. The characteristic roof shape is modelled for the building on the basis of the building object of the cadastre and its roof shape attribute, provided that defined recording criteria are met. However, it may sometimes be necessary to model the roof manually.
U\-sing cadastre data as the basis for modeling, the 3D building models also obtain their positional accuracy. The accuracy of the coordinates from the cadastre varies, as the creation of the coordinates varies~\cite{alkis.2025}. Therefore, no standardised value can be given for the positional accuracy. On average, according to an empirical check, the sampled building corners are within \SI{10}{\cm} of the airborne-image-based 3D mesh. The height accuracy can be given as approx. \SI{1}{\m}.

\begin{table}
\begin{tabular}{lp{5.4cm}}
Standard & CityGML, Level of Detail 2 (LoD2) \\
Geometry & Building footprints with prismatic roofs (generalised) \\
Source data & Cadastral outlines; 5\,m DTM; LiDAR / image-matching point clouds \\
Horizontal ref. & Inherited from cadastre (city mapping frame) \\
Vertical ref. & Derived from point-cloud terrain heights \\
Empirical check & Most sampled building corners within $\sim$10\,cm of mesh \\
\end{tabular}
\end{table}
\begin{figure}
    \centering
    \includegraphics[width=0.5\linewidth]{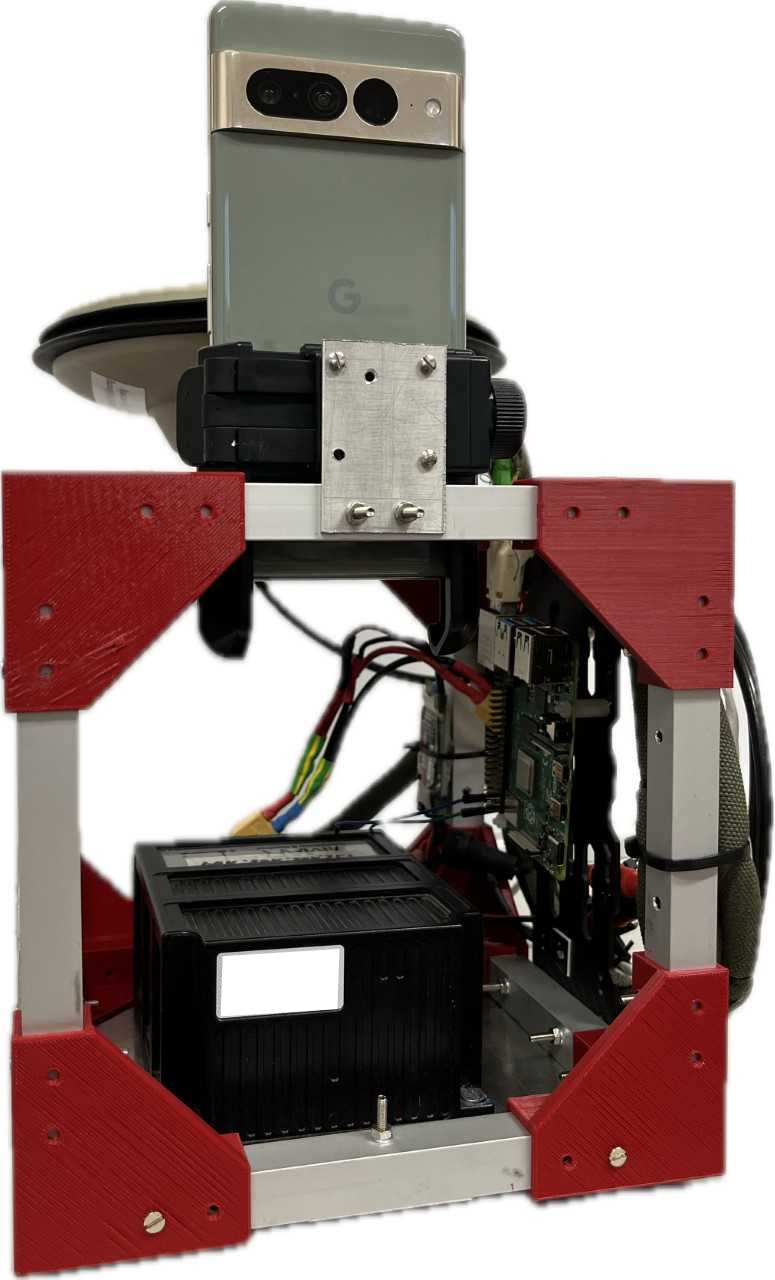}
    \caption{Smartphone rigidly mounted to INS system to capture query data.}
    \label{fig:smartphone_rig}
\end{figure}

\subsection{Query dataset}
Ground-level query images were captured in January 2024 with a handheld smartphone rigidly mounted to a tactical-grade INS (Fig.~\ref{fig:smartphone_rig}). Images were resampled to \si{960 x 1280}{px}, yielding an average GSD of $\sim$4\,cm. See Fig.~\ref{fig:data_overview_query}. 

A precise trajectory was estimated via Post-Processed Ki\-ne\-ma\-tics (PPK) and refined with a Structure-from-Motion and bundle adjustment using Ground Control Points (GCPs) and Check Points (CPs) measured with RTK GNSS.

\begin{tabular}{lp{5.4cm}}
GCP RMSE & (4, 4, 7)\,cm in X, Y, Z \\
CP mean error & (10, 10, 8)\,cm in X, Y, Z \\
Image pose mean error & 8\,mm (X,Y), 3\,mm (Z) \\
Image pose std. dev. & 7\,mm (X,Y), 1\,mm (Z) \\
Orientation mean error & $0.04^\circ$ ($\sigma = 0.03^\circ$) \\
\end{tabular}

The collected dataset comprises 2709 RGB images. To generate the query set, we estimate a subset of non-co-visible images by first rendering the complete set of views using the mesh data. The depth data are then used to compute the full co-visibility matrix between all image pairs. To identify the largest non-co-visible subset, the matrix is converted into a graph where image pairs with fewer than 10\% co-visible pixels are considered non-co-visible. Finally, we compute the maximum independent set on this graph using the solver proposed by \cite{kamis}.
This set of non co-visible images has been further refined through manual inspection to filter out any spuriously selected image resulting from poor meshing. The final released dataset consists of a test set  comprised of 42 non-co-visible images for which the ground-truth position is withheld, and a validation set of 412 consecutive images for which the ground truth poses are provided.



\section{Benchmark Baseline}

To establish a baseline for this dataset, we evaluate MeshLoc \cite{panekMeshLocMeshBasedVisual2022a} and our previous Visual Localisation method \cite{vultaggio_et_al_lc3d2024}. MeshLoc assumes that suitable poses for rendering the reference views correspond to those from which the images used to generate the map were originally captured. In our case, this would imply rendering views from an altitude of 1550 m (i.e. acquisition altitude of the aerial dataset); therefore, we instead employ the initialisation procedure described in our earlier work.

To initialise both methods, we render views in the mesh using open street map data to extract the road network and then sample views along the path. In order to compare both methods in a fair way, we use the same retrieval procedure for both methods where we first retrieve the 500 most visually similar images from the pre-rendered views using CosPlace~\cite{cosplace} global descriptors tuned to work on synthetic images~\cite{bertonMeshVPRCitywideVisual2024a} and then filter them to obtain the 50 closest ones to the smartphone's GNSS pose estimate. Subsequently, the query image is matched against the subset of promising images and the full set of 2D-3D correspondences is used to compute the final pose estimate. To have more details on the initialisation procedure, please refer to our earlier work \cite{vultaggio_et_al_lc3d2024}. Both methods use SuperPoint \cite{detoneSuperPointSelfSupervisedInterest2018b} to extract local features.

\begin{table}[h]
\centering

\begin{tabular}{@{}cccc@{}}
\toprule

Method  & 0.5m,2°/2m,5°/5m,10° $\uparrow$  & ME $\downarrow$      & Time  $\downarrow$  \\ 
 & \textbf{\% / \% / \%} & m , °  & s \\ \midrule
MeshLoc & \textbf{19.05} / 66.67 / 76.19 & 0.97 , 0.63 & 290 \\
Ours    & \textbf{19.05} / \textbf{69.05} /\textbf{ 78.57} & \textbf{0.89} , \textbf{0.56} & \textbf{41}\\ \bottomrule
\end{tabular}%

\caption{Results of MeshLoc and our method on the \AVL \/ dataset. We report the percentage of the 42 query frames localised within each threshold and the median translation and rotational error.}
\label{tab:baseline table}
\end{table}

The results are presented in table \ref{tab:baseline table} and show the percentage of frames localised within three thresholds in accordance with previous visual localisation literature \cite{aachen11,vultaggio_et_al_lc3d2024} and the median translation and rotation error over all the frames. We also report the execution time of both methods, although this will not be assessed in the final rankings, as not all methods will be run on the same hardware. 

The results show that when using the same retrieved images, both methods perform equally well. Our earlier method performs slightly better due to the use of a more advanced pose estimation and refinement pipeline which includes a modern RANSAC \cite{barathMAGSACFastReliable2020} and PnP~\cite{vultaggioPerspectivenPointPracticePerformance2025}. It is also apparent that this dataset is more challenging than any other available benchmark dataset present in the literature when comparing the accuracy achieved by MeshLoc on our dataset. For instance, employing MeshLoc on the Aachen dataset~\cite{aachen11}, it becomes obvious that the performance has degraded as a result of the more challenging texture and possibly the automatic reference pose generation process.

\section{Timeline}

The following schedule is envisaged:
\begin{itemize}
    \item November 2025: Release of data
    \item May 2026: deadline for submissions of solutions, coming along with a short description of the used method, if it should be included in the paper/workshop.
\end{itemize}

\section{Conclusion}

We introduce \AVL, a benchmark for mesh- and object-based visual localisation that combines an airborne-image-derived city mesh with a CityGML LoD2 model and centimetre-accurate, map-independent ground truth for smartphone queries. By constructing a non-co-visible query subset via a maximum independent set on a co-visibility graph, the benchmark enforces true cold-start localisation while the sequential validation split supports training and self-validation. Crucially, by relying on deployable and standardised geospatial assets rather than dense, photorealistic SfM maps, \AVL\ is designed to challenge truly scalable approaches - those that can operate with modest storage/compute footprints and be maintained at city or national scale. A public leaderboard with binning metrics across pose-error thresholds and global statistics ensures rigorous, like-for-like comparison across reference types.

Our baseline evaluations with MeshLoc and our approach are designed to make results directly comparable to the prevailing mesh-based literature. Using identical retrieval procedure and the same evaluation protocol, both methods achieve similar accuracy, while our approach attains substantially lower runtime. We emphasise that our method is early-stage and primarily included to establish a clear point of reference; speed is reported for context rather than ranking. 

We hope \AVL\ will catalyse research on (i) viewpoint selection and rendering pose design for aerial-to-ground matching, (ii) robust cross-domain correspondence between real images and texture-less or low-detail geometric models (e.g.\ LoD2), (iii) retrieval and initialisation strategies that exploit weak priors, and (iv) learning-based pose regression that can leverage sparse or simplified 3D. We also encourage work that explicitly optimises storage, rendering, and compute budgets as first-class scalability objectives.

We release data, evaluation code, a public leaderboard, and plan a community event to consolidate progress. Future extensions will broaden geographic coverage and diversify query acquisition conditions (e.g.\ UAV, bicycle) to further stress-test generalisation and to evaluate methods at larger spatial scales.


\section*{ACKNOWLEDGEMENTS}
\begin{figure}[h]
    \centering
    \includegraphics[width=0.5\linewidth]{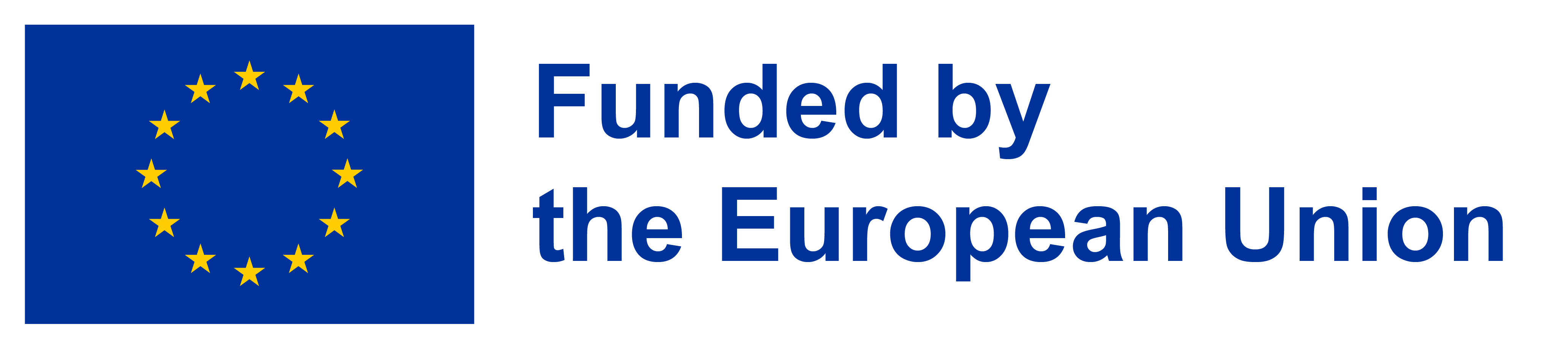}
\end{figure}
This work is part of the EU-Horizon project \textit{egeniouss}\footnote{https://www.egeniouss.eu/}, which received funding under the call HORIZON-EUSPA-2021-Space with the project number 101082128.

{
	\begin{spacing}{1.17}
		\normalsize
		\bibliography{ISPRSguidelines_authors} 
	\end{spacing}
}

\end{document}